LETTER

# Solving Time of Least Square Systems in Sigma-Pi Unit Networks

Pierre Courrieu

Laboratoire de Psychologie Cognitive, UMR CNRS 6146, Université de Provence
29 avenue Robert Schuman, 13621 Aix-en-Provence, France
E-mail: courrieu@up.univ-mrs.fr

*Abstract*– The solving of least square systems is a useful operation in neurocomputational modeling of learning, pattern matching, and pattern recognition. In these last two cases, the solution must be obtained on-line, thus the time required to solve a system in a plausible neural architecture is critical. This paper presents a recurrent network of Sigma-Pi neurons, whose solving time increases at most like the logarithm of the system size, and of its condition number, which provides plausible computation times for biological systems.

## 1. Introduction

Minimizing quadratic error functions is a very usual operation in neurocomputational modeling of learning. This is commonly achieved by applying gradient methods, as in the very popular Error Back-propagation algorithm associated to Multilayer Perceptrons [1,2], but it can also be achieved by using other methods of linear algebra and matrix computation, particularly in the case of Radial Basis Function Networks [3,4]. Learning is not the only context in which one must minimize quadratic error functions: pattern matching operations, that are involved in pattern analysis and pattern recognition models, can also take advantage of the use of effective least square solutions. As an example, assume that patterns are represented as sequences of m points in a real n-dimensional space, or equivalently by real matrices of size $m \times n$, with $m \geq n$. Assume also that patterns are invariant across linear transformations of the form **Y**=**XT**, where **X** is a $m \times n$ pattern, and **T** is a real $n \times n$ transformation matrix. If one must compare an input pattern **X** to a memorized model pattern **M**, then one can basically compute the distance ||**X** - **M**||, however, given the invariance to linear transformations, it is much more relevant to compute the transformation matrix **T** such that the quantity ||**XT** - **M**|| is minimum, because in the case where **X** is a linear transform of **M**, the result will be a zero difference measure, as needed. The matrix **T** is the solution of a least square system, and one knows that

$$\mathbf{T} = (\mathbf{X'X})^{-1}\mathbf{X'M},$$

where the apostrophe (') denotes the transposition operator. Thus, one must compute matrix products, and one must also inverse matrices of the form **X'X**. This requires that **X** be of rank *n*, which is usually the case in practical problems, however, if **X** is not of rank *n*, then the above solution must be replaced by

$$\mathbf{T} = \mathbf{X}^{+}\mathbf{M},$$

where $\mathbf{X}^{+}$ denotes the Moore-Penrose pseudo-inverse of **X** [5]. Unfortunately, to date, one does not know any way of rapidly compute Moore-Penrose pseudo-inverses in neural systems, hence we restrict our approach to the non-singular case. Given that pattern matching is an operation that must be performed on-line, it requires rapid computation. Thus, one can ask whether there are neural architectures, with slow elementary processors (like biological neurons), that could solve least square systems of a given size in a realistic time with respect to known performance of biological systems. Depending on the answer to this question, the use of on-line least square system solving in neural modeling appears to be, *a priori*, a plausible or an implausible hypothesis.

## 2. Solving Process

There are two main types of commonly-used artificial neurons, i.e., distance-based neurons used in RBFNs, and neurons based on a linear combination of input values as in Multilayer Perceptrons. In this last case, each neuron computes a scalar product of a variable input vector with a fixed synaptic weight vector, while synaptic weights can only change slowly during learning processes. However, certain authors hypothesized that neurons could also perform scalar products between variable input vectors, which requires multiplicative connections and leads to the so-called "Sigma-Pi Unit" model [6,7,8], and "Product Unit" model [9,10]. In its general form, the activation of a Sigma-Pi unit whose input is ($z_1, ..., z_k$) is given by

$$f\left( \sum_{S_j \in P_k} w_j \prod_{i \in S_j} z_i \right),$$

where $P_k$ is the set of subsets of the set $\{1, ..., k\}$, $w_j$ is the weight associated by the unit to the $j$th subset (most of these weights are zero), the empty subset $S_0$ has conventionally an associated input product equal to 1, and $f$ is a non-decreasing function. In this paper, we use only $f(x)=x$. This allows the computation of any sum of products, including the usual scalar product of two vectors. A bi-dimensional layer (R) of Sigma-Pi units can simply compute the product (PQ) of two input matrices, since the unit $r_{ij}$ of R must only compute the scalar product of the $i$th row vector of P with the $j$th column vector of Q. The computation of the inverse of a matrix is somewhat more complex, however, in the case of a symmetric positive definite matrix (as **X'X** above), there is an effective solution using a recurrent two-layered network of Sigma-Pi units. We study this solution hereafter. Consider the $n \times n$ symmetric matrix $\mathbf{A} = \alpha \mathbf{X'X}$, for a real $\alpha > 0$. If one knows $\mathbf{V} = \mathbf{A}^{-1}$, then one easily obtains $(\mathbf{X'X})^{-1} = \alpha \mathbf{V}$. Consider a network made of two bi-dimensional layers, **U** and **V**, of $n \times n$ Sigma-Pi units each, while $\mathbf{U}_t$ and $\mathbf{V}_t$ denote the activation states at time $t$ of **U** and **V**, respectively. The layer **V** receives input connections from **U** and itself, while the layer **U** receives input connections from **V** and from a bi-dimensional layer **A** of $n \times n$ cells, whose activation is given by the matrix **A** defined above. Let **I** denote the identity matrix, then one can define the inverting process of **A**, in matrix notation, by

**Definition 1**

$\mathbf{V}_0 = \mathbf{I}$

$\mathbf{U}_{t+1} = 2\mathbf{I} - \mathbf{V}_t \mathbf{A}$

$\mathbf{V}_{t+1} = \mathbf{U}_{t+1} \mathbf{V}_t$

One can immediately see that $\mathbf{V}_t = \mathbf{A}^{-1}$ is a fixed point of the process stated above, hence, we must determine the conditions under which the process converges to this fixed point.

**Theorem 1**

Let **A** be a symmetric, positive definite matrix whose all eigenvalues are in the open interval (0, 2), and let $\omega$ be the eigenvalue of **A** that is the most different from *1*. Then the sequence of matrices ($\mathbf{V}_t$) generated by the process stated in Definition 1 is convergent, and for any real $\varepsilon \in (0, 1)$, one has

$$t \geq \log_2 \frac{\ln \varepsilon}{\ln|1-\omega|} \quad \Rightarrow \quad \|\mathbf{I} - \mathbf{V}_\tau \mathbf{A}\|_2 \leq \varepsilon.$$

*Note*: remember that the matrix norm $\|.\|_2$ is the greatest singular value of the matrix, which is equal to the greatest absolute eigenvalue (spectral radius) if the matrix is symmetric.

*Proof.* Let $\mathbf{B} = \mathbf{I} - \mathbf{A}$, where **A** is invertible by hypothesis. One can write

$$\left( \sum_{i=0}^{k} \mathbf{B}^i \right)(\mathbf{I} - \mathbf{B}) = \mathbf{I} - \mathbf{B}^{k+1} \quad \Leftrightarrow \quad \sum_{i=0}^{k} \mathbf{B}^i = \left(\mathbf{I} - \mathbf{B}^{k+1}\right)(\mathbf{I} - \mathbf{B})^{-1},$$

and thus

$$\sum_{i=0}^{k} (\mathbf{I} - \mathbf{A})^i = \left(\mathbf{I} - (\mathbf{I} - \mathbf{A})^{k+1}\right)\mathbf{A}^{-1} \quad \Leftrightarrow \quad \mathbf{I} - \left(\sum_{i=0}^{k} (\mathbf{I} - \mathbf{A})^i \right)\mathbf{A} = (\mathbf{I} - \mathbf{A})^{k+1}. \tag{1}$$

Since $\|\mathbf{I} - \mathbf{A}\|^2 = |1-\omega| < 1$, one obtains

$$\left\| \mathbf{I} - \left(\sum_{i=0}^{k}(\mathbf{I}-\mathbf{A})^i\right)\mathbf{A} \right\|_2 = \left\| (\mathbf{I}-\mathbf{A})^{k+1} \right\|_2 = |1-\omega|^{k+1}, \tag{2}$$

which converges to 0 as $k$ tends to infinity.

Now, assume that for a given $t$, on can write $\mathbf{V}_t$ in the form

$$\mathbf{V}_t = \sum_{i=0}^{k}(\mathbf{I}-\mathbf{A})^i, \tag{3}$$

then, after the process statement, one has

$$\mathbf{V}_{t+1} = \left(2\mathbf{I} - \left(\sum_{i=0}^{k}(\mathbf{I}-\mathbf{A})^i\right)\mathbf{A}\right)\left(\sum_{i=0}^{k}(\mathbf{I}-\mathbf{A})^i\right).$$

And, after Eq.(1), this is equivalent to

$$\mathbf{V}_{t+1} = \left(\mathbf{I}+(\mathbf{I}-\mathbf{A})^{k+1}\right)\left(\mathbf{I}-(\mathbf{I}-\mathbf{A})^{k+1}\right)\mathbf{A}^{-1} = \left(\mathbf{I}-(\mathbf{I}-\mathbf{A})^{2(k+1)}\right)\mathbf{A}^{-1}.$$

Again after Eq.(1),

$$\mathbf{V}_{t+1} = \sum_{i=0}^{2k+1}(\mathbf{I}-\mathbf{A})^i. \tag{4}$$

On the other hand, since $\mathbf{V}_0 = \mathbf{I}$, one has

$$\mathbf{V}_1 = 2\mathbf{I} - \mathbf{A} = \sum_{i=0}^{1}(\mathbf{I}-\mathbf{A})^i,$$

thus, Eqs. (3) and (4) hold.

In order to express the upper index of the sum in (3) and (4) as a function $K$ of $t$, we note that

$$K(t+1) = 2K(t)+1, \quad \text{and} \quad K(0) = 0,$$

which means that $K(t) = 2^t - 1$, thus:

$$\mathbf{V}_t = \sum_{i=0}^{2^t-1}(\mathbf{I}-\mathbf{A})^i. \tag{5}$$

Substituting Eq.(5) in Eq.(2), one obtains:

$$\|\mathbf{I} - \mathbf{V}_t\mathbf{A}\|_2 = |1-\omega|^{2^t},$$

and finally

$$|1-\omega|^{2^t} \leq \varepsilon \quad \Leftrightarrow \quad t \geq \log_2 \frac{\ln \varepsilon}{\ln|1-\omega|},$$

which completes the proof. □

## 3. Convergence Speed

Theorem 1 states that the inverting process (Definition 1) is convergent provided that all eigenvalues of the matrix $\mathbf{A} = \alpha\mathbf{X'X}$ are in the open interval (0, 2). Assuming that $\mathbf{X'X}$ is not singular, this can be achieved by a suitable choice of the scale factor $\alpha > 0$. There is theoretically an optimal $\alpha$, hereafter named $\alpha_0$, that provides a minimum solving time independent of the size of the system matrix. Let $\{\lambda_i; i = 1,..n, \lambda_j \leq \lambda_{j+1}\}$ be the set of eigenvalues of $\mathbf{X'X}$. The condition number (cond$_2$) of this matrix is $\kappa(\mathbf{X'X}) = \lambda_n / \lambda_1$. According to Theorem 1, one must find a positive real $\alpha$ such that all eigenvalues of $\mathbf{A} = \alpha\mathbf{X'X}$ are in (0, 2), and $|1-\omega|$ is minimum. Thus, an optimal $\alpha$ must satisfy $2 - \alpha\lambda_n = \alpha\lambda_1$, that is

$$\alpha_0 = \frac{2}{\lambda_n + \lambda_1} \tag{6}$$

**Theorem 2**

With $\mathbf{A} = \alpha \mathbf{X'X}$, whenever $\kappa(\mathbf{X'X})$ tends to infinity, the number of iterations of the process stated in Definition 1 is about $\log_2|\ln\varepsilon| + \log_2(1+\kappa(\mathbf{X'X}))-1$.

*Proof.* Taking the result of Theorem 1, and given that both $\varepsilon$ and $|1-\omega|$ are in (0, 1), one can write

$$\log_2\frac{\ln\varepsilon}{\ln|1-\omega|} = \log_2|\ln\varepsilon| - \log_2|\ln|1-\omega||. \tag{7}$$

On the other hand, using the scale factor $\alpha_0$ defined in Eq.(6), one obtains $\omega = \frac{2\lambda_1}{\lambda_n + \lambda_1}$, which tends to zero whenever $\kappa(\mathbf{X'X})$ tends to infinity. By the equivalence $\ln(1+x) \approx x, x \to 0$, one obtains

$$|\ln|1-\omega|| \approx \frac{2\lambda_1}{\lambda_n + \lambda_1} \quad ,$$

and by substitution in Eq.(7)

$$\log_2|\ln\varepsilon| - \log_2\frac{2\lambda_1}{\lambda_n + \lambda_1} = \log_2|\ln\varepsilon| + \log_2\frac{\kappa(\mathbf{X'X})+1}{2} \quad ,$$

which completes the proof. $\square$

It is clear that one cannot compute $\alpha_0$ without knowing the two extreme eigenvalues of $\mathbf{X'X}$, except if $n=2$ since the trace of a matrix (sum of its diagonal coefficients) is equal to the sum of its eigenvalues, and thus, for $n=2$, it suffices to divide 2 by the trace in order to obtain $\alpha_0$. Nevertheless, one can generalize this method for $n>2$, and the obtained scale factor is lower than $\alpha_0$, thus it is not optimal, however it is suitable since it guarantees that all eigenvalues of the rescaled matrix are in (0, 2), thus the inverting process converges. We set

$$\alpha_1 = \frac{2}{trace(\mathbf{X'X})}. \tag{8}$$

**Theorem 3**

With the $n \times n$ matrix $\mathbf{A} = \alpha_1 \mathbf{X'X}$, for $n \geq 2$, whenever $\kappa(\mathbf{X'X})$ tends to infinity, the number of iterations of the process stated in Definition 1 is at most $\log_2|\ln\varepsilon| + \log_2(\kappa(\mathbf{X'X})) + \log_2 n - 1$.

*Proof.* After Eq.(8), one has $\alpha_1 \geq \frac{2}{n\lambda_n}$, which implies that $\omega = \alpha_1\lambda_1 \geq \frac{2\lambda_1}{n\lambda_n} = \frac{2}{n\kappa(\mathbf{X'X})}$. Substituting this last quantity in Eq.(7), a proof similar to that of Theorem 2 leads to Theorem 3, where the "at most" results from the fact that the substituted term is in fact lower than $\omega$. $\square$

As one can see, the scale factor $\alpha_1$ defined in Eq.(8) leads the number of iterations of the inverting process to depend on the size $n$ in $O(\log n)$. Now, set $\mathbf{Z} = \mathbf{X'X}$, and let us define another scale factor as

$$\alpha_2 = \frac{2}{\min_{1\leq i\leq n} z_{ii} + \max_{1\leq i\leq n}\sum_{j=1}^{n}|z_{ij}|}. \tag{9}$$

**Theorem 4**

If $\mathbf{Z} = \mathbf{X'X}$ is not singular, then the inverting process stated in Definition 1, applied to the matrix $\mathbf{A} = \alpha_2 \mathbf{Z}$, is convergent.

*Proof.* In order to prove this, it suffices to prove that $\alpha_2 \leq \alpha_0$. After Gershgorin-Hadamard's theorem, one has

$$\lambda_n \leq \max_{1 \leq i \leq n} \sum_{j=1}^n |z_{ij}| \left(= \|Z\|_\infty\right).$$

On the other hand, since **Z** is a symmetric matrix, it can be written as **Z=QDQ'**, where **Q** is the orthogonal matrix of eigenvectors, and **D** is the diagonal matrix of eigenvalues of **Z**. Then one has:

$$z_{ii} = \sum_{j=1}^n q_{ij}^2 \lambda_j \quad \text{with} \quad \sum_{j=1}^n q_{ij}^2 = 1,$$

which implies that $\lambda_1 \leq \min_{1 \leq i \leq n} z_{ii}$, and thus $\alpha_2 \leq \alpha_0$, which completes the proof. □

The main interest of the scale factor $\alpha_2$ defined in Eq.(9) is that it can approximate the optimum $\alpha_0$ better than $\alpha_1$, particularly in the case of a large system matrix. Unfortunately, we have no analytical expression of the convergence speed of the inverting process as a function of *n* with $\alpha_2$, thus we must study it experimentally.

## 4. Computational Tests

A set of 120 random test matrices was generated, according to Moré and Toraldo's method [11]. Each test matrix was of the form **Z** = **X'X**, with **X** = **D**$^{1/2}$**H**, **D** = $diag(\lambda_1, ..., \lambda_n), \lambda_i = \kappa^{(i-1)/(n-1)}$, and a random Householder's matrix **H** = **I** – 2 **hh'**/**h'h**, with **h** uniformly random in (-1, 1)$^n$. The size *n* of the test matrices was varied from 16 to 256, and the condition number $\kappa$ was varied from 64 to $2^{20}$, using only powers of 2. The inverting process stated in Definition 1 was applied to the test matrices rescaled with each of the three scale factors $\alpha_0$, $\alpha_1$, and $\alpha_2$, defined in Eqs. (6), (8), and (9) respectively, the process ending when all the coefficients of the error matrix ( **I** – **V**$_t$**A** ) were lower than $10^{-6}$ in absolute value. Let $N_j$ denote the number of iterations obtained with the scale factor $\alpha_j$, then the computational results can be summarized as:

$N_0 = \log_2 \kappa + 3 \, (\pm 0),$
$N_1 = \log_2 \kappa + \log_2 n + 1 \, (\pm 0),$
$N_2 = \log_2 \kappa + \frac{1}{3} \log_2 n + 2.433 \, (\pm 0.22).$

As one can see, $N_2$ depends on *n* in $O\left(\log \sqrt[3]{n}\right)$, and $\alpha_2$ provided faster convergence than $\alpha_1$ for $n \geq 5$.

However, it is not convenient to express the number of iterations as a function of the condition number $\kappa$, since this number is not easy to compute in practical cases. One can provide suitable statistics by considering the size ($m \times n$) of the matrix **X**, and making some reasonable hypothesis concerning the distribution of data points.

Patterns can commonly be represented as sequences of m points resulting from an encoding process, and the properties of the resulting matrix **X** of course depend on each particular encoding procedure. As an example, if one hypothesizes that the n encoding dimensions are independent, and that these variables are all uniformly distributed in the same interval of center zero (say [-1, 1]), then the number of iterations of the inverting process (with precision as in the previous test) depends on the size ($m \times n$) of the matrix **X** as reported in Table 1 (where each mean corresponds to 10 distinct random matrices **X**). One can observe that the number of iterations increases as n increases, but it decreases as the ratio *n/m* decreases (because this statistically improves the condition number of **X** '**X**, under the above hypothesis). Now, in order to solve least square systems in a parallel neural architecture, one must add to the inverting process a number of sequentially dependent operations: the product **X'X** (and simultaneously **X'M**), the computation of $\alpha$, the rescaling **A**=$\alpha$**X'X**, the products **T**=($\alpha$**V**)(**X'M**), **XT**, and the distance ‖ **XT** –**M**‖. All these operations are of sigma-pi type, except the last one, that is of Radial Basis Function type [3, 4], and the computation of $\alpha$, that requires a division, and possibly min and max operations. Divisions can be computed by product units [9, 10], which are in fact variants of sigma-pi units using possibly negative exponents. On the other hand, max-like operations have also been hypothesized to be basic neural operations [12]. In summary, we must add about 7 sequentially dependent operations to that of the inverting process (with 2 operations per iteration) in order to complete the comparison (which does not include the encoding process whose output is **X**). As an example, using Table 1 for a pattern of 512 points in R$^8$, with $\alpha = \alpha_2$, one obtains 2 × 4 + 7 = 15 sequentially dependent operations. If one estimates the time per neural operation, in a biological system, to be about 5 milliseconds [13], then the above example requires about 75 milliseconds, which is quite reasonable as compared to human perception performance (e.g. an ocular fixation of about 250 milliseconds is required to recognize a printed familiar word).

**Table 1.** Mean number of iterations ($N_1$ for $\alpha_1$, and $N_2$ for $\alpha_2$) as a function of the size ($m \times n$) of the matrix $X$.

|  | n = 4 | n = 8 | n = 16 | n = 32 | n = 64 |
|---|---|---|---|---|---|
| m = n |  |  |  |  |  |
| $N_1$ | 11.1 (± 3.2) | 13.7 (± 2.8) | 16.8 (± 3.2) | 18.4 (± 2.1) | 24.1 (± 6.1) |
| $N_2$ | 10.9 (± 3.4) | 13.1 (± 3.2) | 15.5 (± 3.1) | 16.4 (± 2.1) | 21.3 (± 6.3) |
| m = 2 n |  |  |  |  |  |
| $N_1$ | 7.8 (± 0.6) | 8.8 (± 0.8) | 10.0 (± 0.5) | 11.4 (± 0.5) | 12.3 (± 0.5) |
| $N_2$ | 7.3 (± 0.9) | 7.7 (± 0.9) | 8.2 (± 0.4) | 9.2 (± 0.4) | 9.0 (± 0.0) |
| m = 4 n |  |  |  |  |  |
| $N_1$ | 6.0 (± 0.5) | 7.6 (± 0.5) | 9.1 (± 0.3) | 10.0 (± 0.0) | 11.0 (± 0.0) |
| $N_2$ | 5.5 (± 0.5) | 6.1 (± 0.7) | 7.0 (± 0.5) | 7.0 (± 0.0) | 7.2 (± 0.4) |
| m = 8 n |  |  |  |  |  |
| $N_1$ | 5.9 (± 0.3) | 7.0 (± 0.0) | 8.0 (± 0.0) | 9.0 (± 0.0) | 10.0 (± 0.0) |
| $N_2$ | 5.0 (± 0.0) | 5.2 (± 0.6) | 5.8 (± 0.4) | 6.0 (± 0.0) | 6.1 (± 0.3) |
| m = 16 n |  |  |  |  |  |
| $N_1$ | 5.2 (± 0.4) | 7.0 (± 0.0) | 8.0 (± 0.0) | 9.0 (± 0.0) | 10.0 (± 0.0) |
| $N_2$ | 4.2 (± 0.4) | 5.0 (± 0.0) | 5.0 (± 0.0) | 5.0 (± 0.0) | 6.0 (± 0.0) |
| m = 32 n |  |  |  |  |  |
| $N_1$ | 5.1 (± 0.3) | 6.3 (± 0.5) | 7.9 (± 0.3) | 9.0 (± 0.0) | 10.0 (± 0.0) |
| $N_2$ | 4.0 (± 0.5) | 4.1 (± 0.3) | 4.4 (± 0.5) | 5.0 (± 0.0) | 5.0 (± 0.0) |
| m = 64 n |  |  |  |  |  |
| $N_1$ | 5.0 (± 0.0) | 6.0 (± 0.0) | 7.0 (± 0.0) | 8.1 (± 0.3) | 9.0 (± 0.0) |
| $N_2$ | 3.5 (± 0.5) | 4.0 (± 0.0) | 4.0 (± 0.0) | 4.0 (± 0.0) | 4.6 (± 0.5) |

**Acknowledgment.** This work was partially supported by a grant from Conseil Général 13, Conseil Régional PACA, and CNRS-Université de Provence.

**Pierre Courrieu** is a CNRS researcher currently working at the University of Provence (France). He published works on visual word recognition, neural computation, data encoding, and global optimization.